# Learning to Detect Vehicles by Clustering Appearance Patterns

Eshed Ohn-Bar, *Member, IEEE*, and Mohan Manubhai Trivedi, *Fellow, IEEE*

***Abstract*—This paper studies efficient means for dealing with intra-category diversity in object detection. Strategies for occlusion and orientation handling are explored by learning an ensemble of detection models from visual and geometrical clusters of object instances. An AdaBoost detection scheme is employed with pixel lookup features for fast detection. The analysis provides insight into the design of a robust vehicle detection system, showing promise in terms of detection performance and orientation estimation accuracy.**

## I. Introduction

Efficient object detection requires robustness to the appearance variations of an object. In the context of vehicle detection studied in this work, these variations may stem from a changing observation angle, illumination variability, vehicle shape and type, truncation out of the camera view, different occlusion levels, etc. Due to their commonality, the handling of such challenges is key to the monitoring of the on-road environment using vision systems.

A main question dealt with in this work is what is the best approach for clustering the training data in order to learn the subcategory/component models. For instance, the successful deformable parts model (DPM) [1], [2] learns a multi-component mixture model using aspect ratio features. Although other elements of the DPM (i.e. deformable parts in a pictorial structure and latent discriminative learning) offer additional robustness and generalization capabilities, an emphasis on the model components is well motivated. Multiple components provide a natural accommodation of object appearance variation due to geometry, orientation, and occlusion. In specific object detection domains, such as vehicle or pedestrian detection, certain appearance patterns (e.g. certain occlusion types) may be common, thereby motivating learning models that are specialized for such well defined patterns. Consequently, the specialized models are learned over more visually homogeneous samples which simplifies the learning task and translates to improved detection performance in test time. Furthermore, several recent studies show components to be useful in varying domains of object detection [2]–[5].

Because the relationship between the clustering step and detection performance is not clear, the preferable approach for obtaining the subcategory component clusters is not trivial, yet it is of great interest for researchers. For instance, clustering can be done using aspect ratio of bounding boxes or visual cues [4]. Common solutions employ a discriminative clustering process (e.g. latent SVM) which may produce degenerate or noisy category clusters [6]. Generally, there is also a trade-off between detection with less models but at greater run time speed vs. detection with more models at better detection performance. This further motivates our study, as subcategory models are expensive to evaluate in test time. Therefore, learning better subcategory models could result in speed gains without hindering detection performance. This aspect of the detection scheme is of particular interest in mobile settings of intelligent transportation systems, where fast and lightweight computation is desired.

In this work, we study object subcategorization using clustering of 3D orientation, position, occlusion level and type, and other geometrical shape features. This study demonstrates the following:

**Features for clustering of object subcategories**: Learning good subcategory models is shown to be highly dependent on the features used for the subcategorization. In particular, it is shown to work best (i.e. produce homogeneous clusters useful for detection) when using a set of 3D geometrical features. Furthermore, clustering techniques are studied and compared in terms of impact on detection performance. The following questions motivate the study of this paper: How should one quantize the data best in order to obtain good subcategory models? How to efficiently employ such a framework to handle occlusion and orientation variation? Should model learning occur over varying occlusion levels, if so, how should these be chosen? Should statistics of the occluder and different occlusion types be considered? What if no 3D orientation information is available? How does the choice of subcategories affect orientation estimation? The novel comparative study in this work provides a step towards answering such questions.

**Multiple components of a fast detection scheme**: Generally, top performing vehicle detectors (e.g. DPM-based approaches) contain several speed bottlenecks. At the same time, pedestrian detection techniques has seen considerable speedups [7], [8]. The work in this paper can be seen as an attempt to adapt the fast approach of [7] to multi-component settings (which is often ignored in pedestrian detection by learning only one rigid template). This adaption results in both good detection performance and fast run time even with many subcategory models. Furthermore, the simpler rigid template detection models are shown to perform significantly better detection than more complicated models which explicitly incorporate the notion of object parts. The approach provides a tuning parameter which controls the trade-off between speed and performance. Depending on the number of subcategories, the entire detection pipeline can run between 13 (for 1 model) to 5 (for 20 models) frames per second (fps) on full resolution images of size 1242 $\times$ 375, and further speedups are possible [9], [10]. Fast detection and recognition of objects is essential

The authors are with the Laboratory for Intelligent and Safe Automobiles (LISA), University of California San Diego, San Diego, CA 92093-0434 USA (e-mail: eohnbar@ucsd.edu; mtrivedi@ucsd.edu).

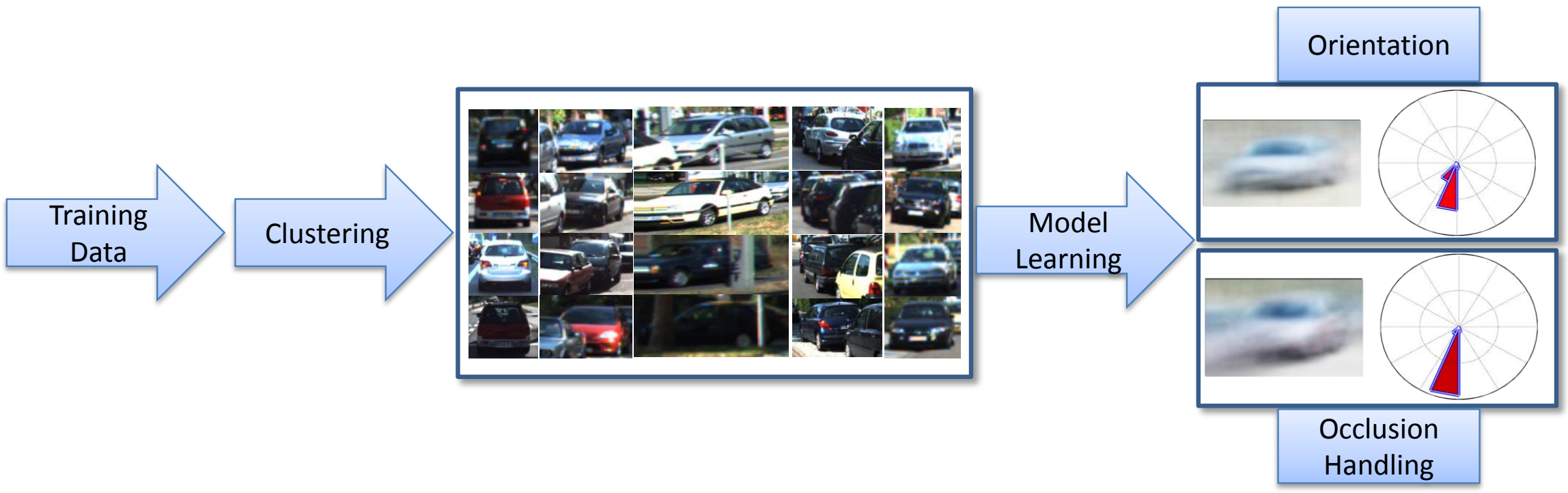

(a) Training pipeline.

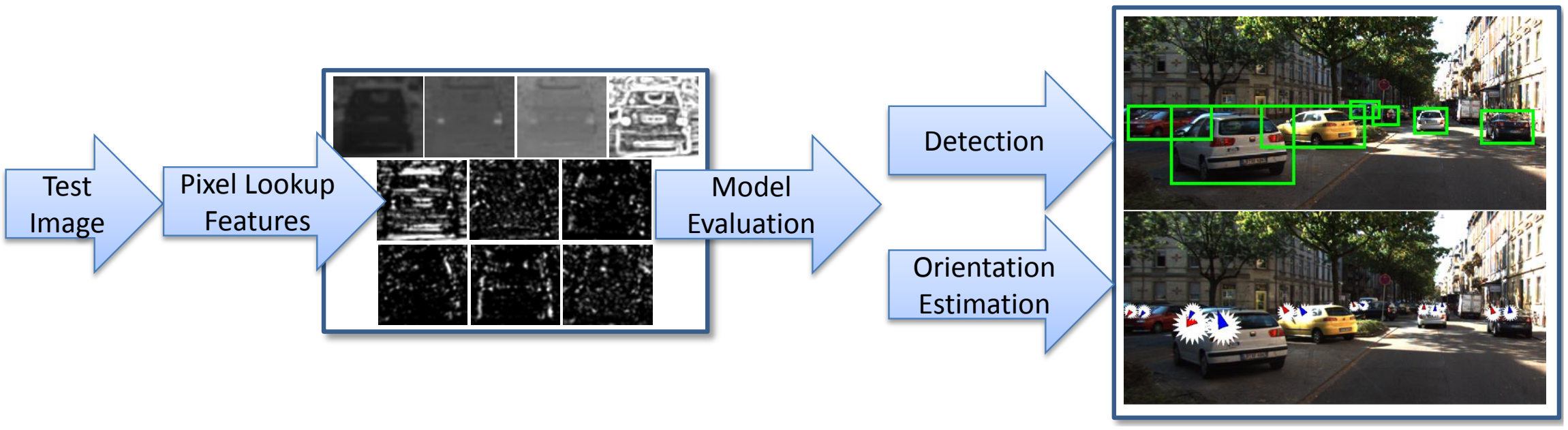

(b) Testing pipeline.

Fig. 1: This work studies different features for learning appearance patterns clusters from a training set. Specifically, the relevance of subcategorization to vehicle detection and orientation estimation is studied. The underlying detector of AdaBoost with color and gradient-based pixel lookup features [7] provides fast detection in test time.

in development of intelligent vehicles applications, for instance in trajectory understanding [11], [12].

**Orientation estimation**: After optimization of the features used for clustering, clustering techniques, and number of clusters, the final detector is used to perform highly robust orientation estimation. This task summarizes the effectiveness of the framework.

## II. Related Research Studies

Commonly, sliding window-based vehicle detection may employ a variant of HOG+SVM (histogram of oriented gradients and a linear support vector machine) [19] or cascade detectors [7]. We review recent relevant literature, and turn the reader to the comprehensive review of [20] for additional detail.

**Vehicle detection with DPM and its variants**: The DPM model [1], [21], which builds on HOG features and a Latent SVM, has also been a common choice for vehicle detection [14], [22]. In [15], a variant of the DPM framework is used in order to detect vehicles under heavy occlusion and clutter. In [22], integrating scene information was shown to improve both the detection and orientation estimation performance of the DPM. To better handle detection of occluded vehicles, a second-layer conditional random field (CRF) was used over root and part score configurations provided by a DPM model in [23]. More recently, modeling structure of part configurations and component models using an AND-OR structure [24] favorably compared against the classical DPM. In the aim of detecting objects under occlusion, a joint object detector was proposed in [25], and bounding boxes were predicted using linear regression. Although the approach in [25] appears promising for pedestrian detection, the study of [5] showed a joint vehicle detector with bounding box regression to perform worse than a single object DPM detector. Nonetheless, a main improvement over the DPM baseline was gained by incorporating mixture components for occluded vehicle cases. This provides further motivation for our study of learning subcategory models for appearance variations In [5], [25], [26], an arbitrary partition of the data is performed to produce an initialization to the LSVM-based assignment. These studies generally consider a subset of the geometrical features studied in this work. Furthermore, an extensive analysis of the choice of subcategorization features is absent from the above studies. Interestingly, in this study initialization of a discriminative clustering of visual data framework (e.g. LSVM) with different geometrical features resulted in only minor improvements. As a matter of fact, only working in the 3D geometry space produced the best results in terms of detection performance. To emphasize, unlike the aforementioned studies, we also study the impact of different features and clustering techniques on the final detection.

**Subcategory learning**: A common approach for improving model generalization is by learning subcategories within an object class. For instance, these are used in conjunction with DPMs in order to detect objects at varying aspect ratios. In [13], visual subcategories corresponding to vehicle orientation were learned in an unsupervised manner using Locally Linear Embedding and HOG features. In [27], an exemplar SVM is learned for each positive example, and the learned weights are used in affinity propagation to generate visual subcategories. This exemplar-based step provides the initialization to LSVM clustering. Several other recent studies have shown the benefit of visual homogeneity in training data on model performance [28], [29]. Vehicle orientation estimation is studied using supervised, semi-supervised, and unsupervised settings with DPM framework in [30], with supervised settings showing the best results.

TABLE I: Overview of related research studies for vehicle detection at multiple orientations and occlusion levels.

| Study | Features | Classifier | Subcategory Clustering | Subcategory features | Parts | Occlusion-Handling | Speed (fps) |
|---|---|---|---|---|---|---|---|
| Kuo and Nevatia [13] (2009) | HOG | GentleBoost | LLE | HOG | Y | N | - |
| Niknejad *et al.* [14] (2012) | HOG | LSVM | LSVM | Aspect-ratio | Y | N | ∼0.5 (640 × 480) |
| Hejrati and Ramanan [15] (2012) | HOG | LSVM | k-means/EM | Part configuration and occlusion type | Y | Y | 0.03 (1242 × 375) † |
| Pepik *et al.* [5] (2013) | HOG | LSVM | Rule-based | 3D orientation and occlusion types | Y | Y | 0.1 (1242 × 375) |
| Li *et al.* [16], [17] (2013) | HOG | AND-OR structure | AND-OR tree | Aspect-ratio and occlusion | Y | Y | 0.3 (1242 × 375) |
| Sivaraman and Trivedi [18] (2013) | Haar | AdaBoost | - | - | Y | Y | 14.5 (500 × 312) |
| This study | Color, gradient orientation, and magnitude | AdaBoost | k-means, spectral clustering, weak/full supervision | Geometrical and visual features | N | Y | 5 (1242 × 375) ‡ |

†: Not reported in the paper but obtained using the publicly available code on a 6 core, Intel Core i7 @ 3.30 GHz 16 GB RAM machine.
‡: Run-time depends on the number of subcategories. This number is for 20.
Index: **LSVM**: Latent Support Vector Machine. **EM**: Expectation Maximization. **HOG**: Histogram of Oriented Gradients.
**fps**: Frames per Second. **LLE**: Locally Linear Embedding.

Discriminative subcategorization, where the clustering considers negative instances, was studied in [6]. The technique is shown to improve results over Latent SVM-based clustering refinement, which is shown to be prone to cluster degeneration. Furthermore, the framework in [6] provided more visually consistent clusters. This technique will be used as a baseline in this work.

In [4], the importance of efficient learning of visual subcategories for different objects is highlighted. By using an extension of the Latent SVM framework initialized with k-means on visual data, a significant gain in performance was shown on the PASCAL dataset. The authors in [4] motivate visual subcategorization over other forms of data partitioning by arguing that tighter clusters can be extracted from visual data, as semantic (human-based) subcategories simply aim to encode visual consistency.

One important advantage of such an approach is that no annotation is required. While this may be correct, we argue that this may substitute a hard problem (of annotation) with another hard problem of visual subcategorization. Furthermore, despite the motivation for subcategorization using visual features, no direct comparison of the impact that different features have on clustering was performed in the aforementioned works. In this paper, we use the KITTI dataset [31] to answer such questions, and visual subcategorization is shown to be significantly inferior to geometrical 3D orientation and occlusion features.

Table I outlines the differences between existing approaches and ours. We pursue an alternative approach to the DPM using a rigid template Viola-Jones style detector [7], [18]. This work focuses on static appearance cue detectors, but occlusion and truncation handling can also be done using dynamic motion cues [32], [33]. Note that the reported run times in Table I vary based on image size and other parameters. For instance, in [14] a limit is set on the minimum size of the detcted objects which could allow for a speedup. For the framework studied in this work, the main parameter is the number of subcategories. Since each subcategory requires a separate model evaluation, the trade-off between speed and performance will be analyzed in Section VII.

## III. Object Subcategorization

The key components of the proposed framework are shown in Fig. 1. Ultimately, the goal is to cluster the training data into visually homogeneous clusters. The smaller intra-cluster ambiguity in turn produces better detection models as opposed to learning one model over all instances (a **monolithic** classifier). A clustering algorithm is used to produce a predetermined number of clusters. As the features for categorization have a great impact on the resulting trained detection models, a variety of visual and fine-grained 3D geometrical features were studied.

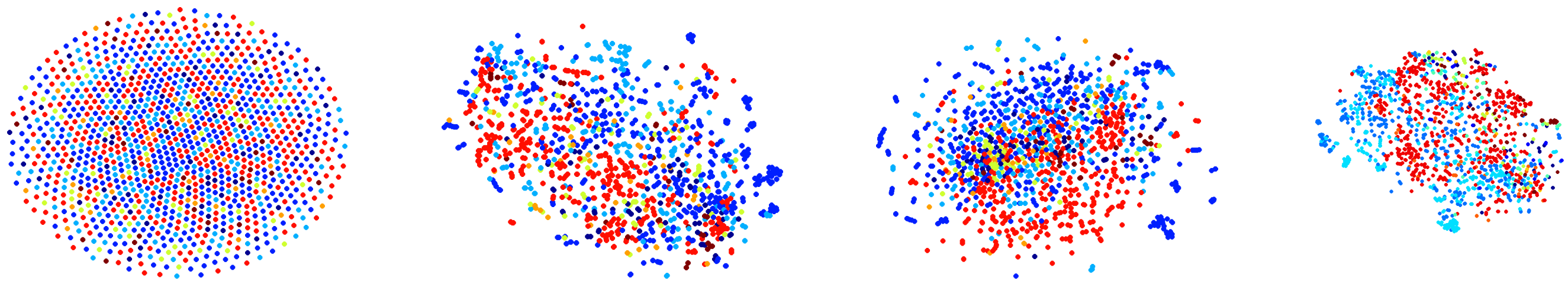

(a) RGB-Color (b) LUV-Color (c) Gradient (d) CNN

Fig. 2: Visualization of the feature space by projection to 2D using t-SNE [34] with varying features on the entire KITTI training dataset (some samples were removed for visualization). The color of each point corresponds to an assigned bin according to annotated vehicle orientation. Clear separation in the feature space translated to better performing detection models in our experiments.

### *A. Visual Clustering*

As shown in Table I, shape features are commonly used for capturing the shape patterns of vehicle instances. Since orientation accounts for much of the appearance variation, Fig. 2 visualizes how well do different types of features capture orientation information. Pixel values in LUV color space are shown to be more useful when compared to the RGB color space. Color may capture cues such as taillights. In the majority of the experiments, a total of 10 types of features are used for visual subcategorization: LUV, normalized gradient magnitude, and oriented gradients at 6 bins (as in [7]). These 10 feature types were all shown useful for detection of vehicles in our experiments, and can be extracted at more than 55 fps on a CPU for full resolution images of size $1242 \times 375$. As pre-processing, all positive instances are resized to the mean image size $[w_m, h_m]$ for clustering. Clustering using the aforementioned fast features is also studied against 4096-D high quality convolutional neural network (CNN) features from Caffe [35], fine-tuned on the PASCAL dataset [36].

### *B. Mining Object Geometry, Orientation, and Occlusion Patterns*

3D object information can be extracted from the scene using a variety of methods and sensors. In the KITTI dataset, images and information from a Velodyne lidar were annotated with 3D bounding boxes. Below, we detail the types of geometry features that were studied in this work, and consequently the different strategies that will be employed to produce clusters using the features.

The availability of high quality 3D information raises the following research question: can these be used to learn detection models, as opposed to visual features? How should these different modalities be integrated to produce the best models? These questions will be studied in this work. To represent vehicle instances, we extract the following set of geometrical features.

**3D orientation**: When detecting vehicles in different driving settings (intersection, highway, etc.), appearance variation due to the observation angle is common. Instead of using the available raw 3D yaw angle (rotation around the Y-axis in camera coordinates), the observation angle is used (relative orientation of the object with respect to the camera) by considering the angle of the vector joining the camera center in 3D and an object. The reason for this is that the yaw angle as it is does not take into account the ego-vehicle, which may be observing the object from different angles. For instance, an object at 90 degrees may appear very differently depending on where it is located around the ego-vehicle.

**Aspect-ratio**: The 2D bounding box of objects is correlated with the geometry of the object being detected. We explicitly include this in the clustering, with the aim of creating a different model for objects at different aspect ratios. These may not necessarily involve different orientations (e.g. a car vs. a truck). Learning models at different aspect ratios provides a significant improvement in detection (as opposed to keep a fixed dimension model for all clusters).

To encode variation in appearance due to occlusion, it may not be necessary to train a model for very fine-grained occlusion levels. Nonetheless, as occlusion level is a main factor in visual diversity, we found that using explicit occlusion features in the clustering can improve detection of partially-occluded vehicles.

**Truncation level**: The percentage of the vehicle outside of the camera view is also used as a feature.

**Occlusion level**: Given a 2D bounding box, we search the 3D space for an occluder, which is the closest vehicle in 3D that is also closer to the camera than the occluded vehicle. The occlusion level feature is calculated as $\frac{area(BB_{occluder} \cap BB_{occludee})}{area(BB_{occludee})}$.

**Occlusion type features**: Since the above process may miss some occlusion information due to unannotated objects, we also use an occlusion index. The index represents whether an object is not occluded, partially occluded, heavily occluded, or includes an unknown occlusion type.

As occluded objects are common in the dataset, a set of features is extracted to further refine the occlusion types. The relative orientation, $\theta_{occludee} - \theta_{occluder}$ provides additional context for the type of the occlusion. For instance, certain patterns are common for certain occluder-occludee orientation combinations. Therefore, the occluder orientation, $\theta_{occluder}$, as well as the relative 3D position, $\mathbf{p}_{occludee} - \mathbf{p}_{occluder}$ ($\mathbf{p} \in \mathbb{R}^3$), are used as features. Finally, a binary feature is used to encode

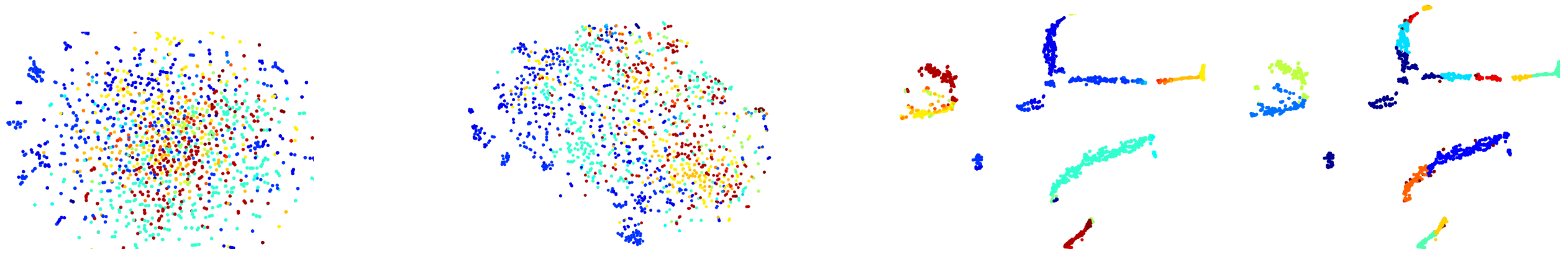

(a) Visual Features (gradient+color) (b) Visual Features (CNN) (c) Geometrical Features (d) Clustering Output

Fig. 3: Visualization of the feature space in 2D (visual features and the proposed set of geometrical features III-B). The color of each point corresponds to a quantization into 10 bins in orientation space and 2 bins in occlusion space (occluded and non-occluded). (d) shows the output of k-means clustering on (c), which may be used for subcategorization.

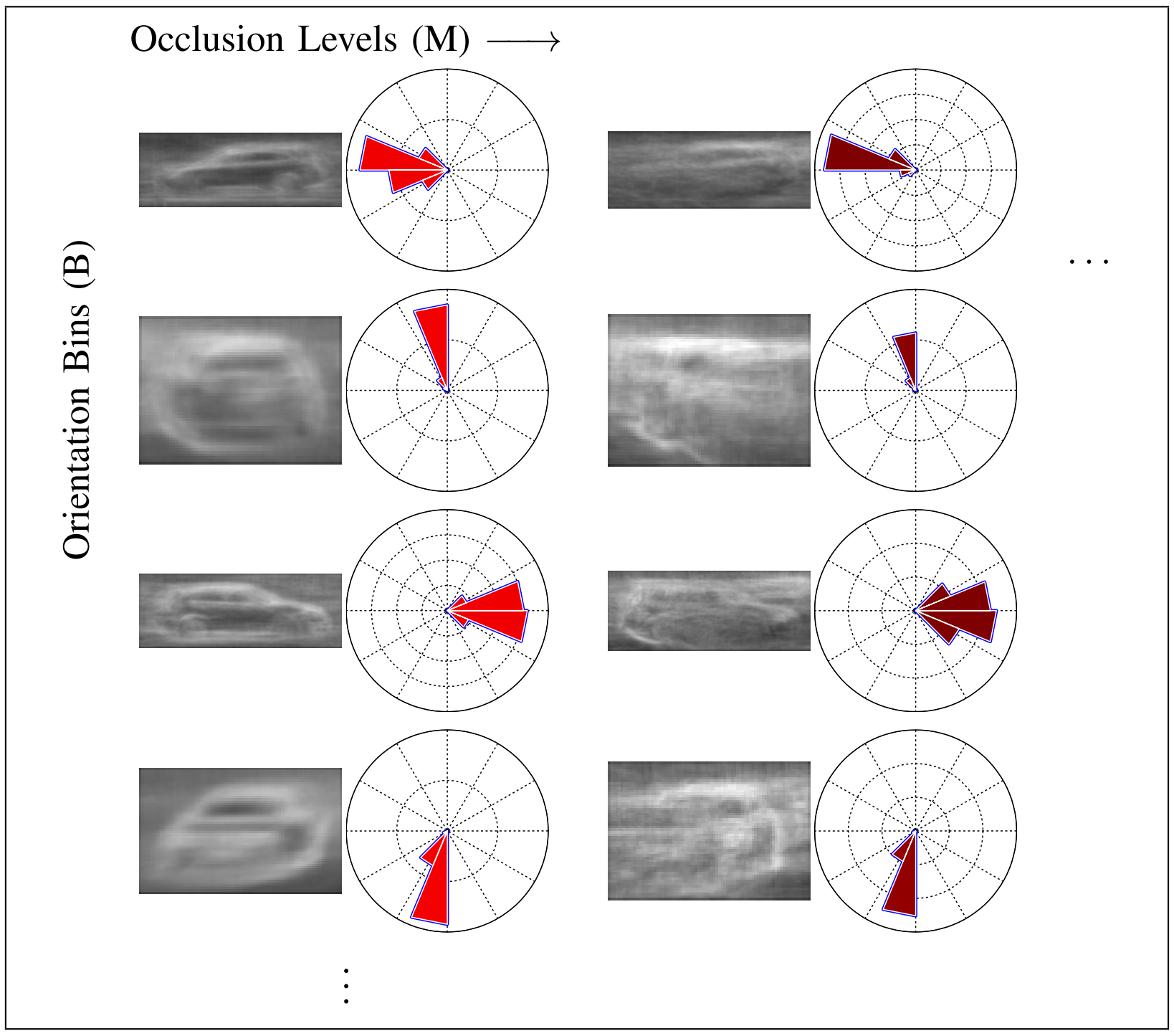

Fig. 4: One subcategorization possibility is to cluster the data according to orientation and occlusion levels. Such strategy would produce the visualized clusters (mean gradient image is shown). Darker rose plots correspond to higher number of occluded samples in the cluster.

whether the occluder is on the left or the right of the occludee using the centroid coordinates of the bounding boxes.

### C. Clustering

Fig. 3 studies the space of categorization features (projected into 2 dimensions using t-SNE [34]). Each point is binned into a color according to its orientation and occlusion level (either occluded or non-occluded). The figure demonstrates how higher quality features (CNN) produce a more clearly separated feature space for clustering. As will be shown in the experimental analysis, this is correlated with improved detection models compared to the gradient+color features. Nonetheless, categorization based on geometrical features produced the best performing detection models in our experiments. Nonetheless, quantization of the geometrical feature space is still not trivial. For instance, certain geometrical variations may not correspond to significant appearance variation (hence such variations can be included in the same model). One possible quantization can be done in a unsupervised or semi-supervised fashion using a clustering algorithm, as shown in Fig. 3.

**Strategy 1**: A uniform binning of orientation bins and occlusion level bins. This is a supervised approach which encodes the prior information that two parameters only (orientation and occlusion level) account for the majority of appearance variations. Even in this straightforward approach, there are several possible quantization techniques.

First, occluded and not occluded vehicles may be grouped together into the same cluster, so that $M$ (the occlusion quantization parameter as shown in Fig. 4) is set to 1 and $B$ is varied. Second, we may entirely split occluded and not occluded cases in all of the analysis, referred to as **Split** in the experimental analysis in Section VII (e.g. quantization over $[0-10\%]$ and $[11-100\%]$). Third, we may vary both $B$ and $M$. For instance, if the maximum occlusion level in all of the samples is 80%, a value of $M=2$ would create a quantization

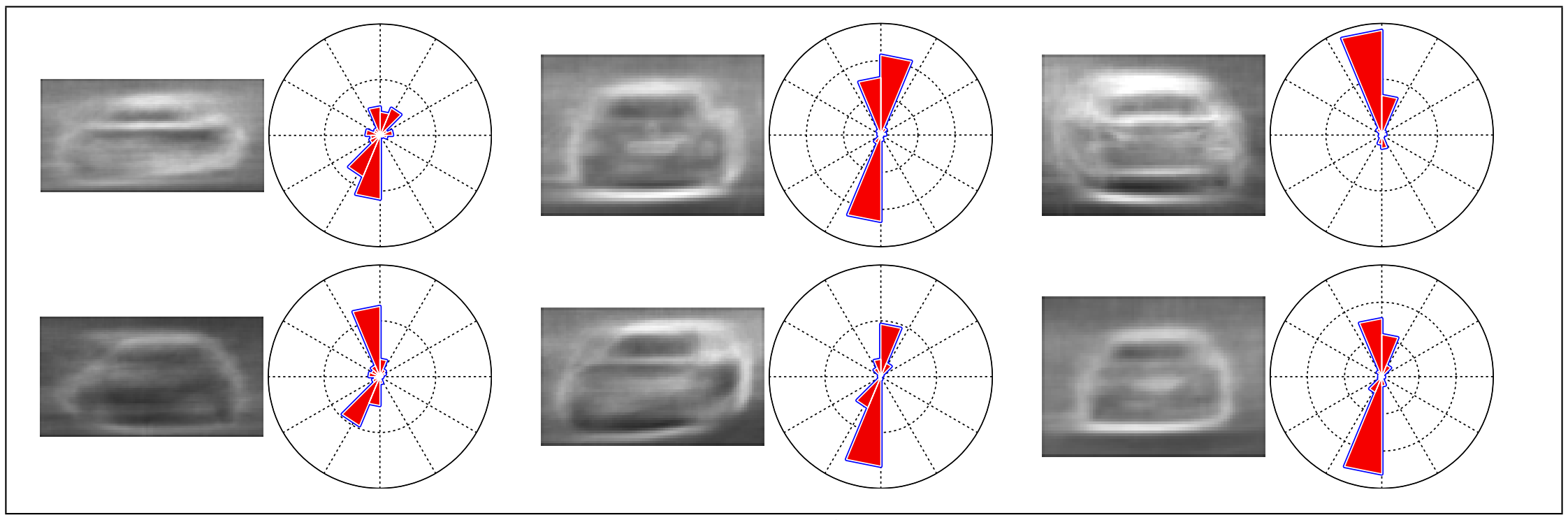

(a) Discriminative subcategorization (**DSC** [6]) with only **visual features**.

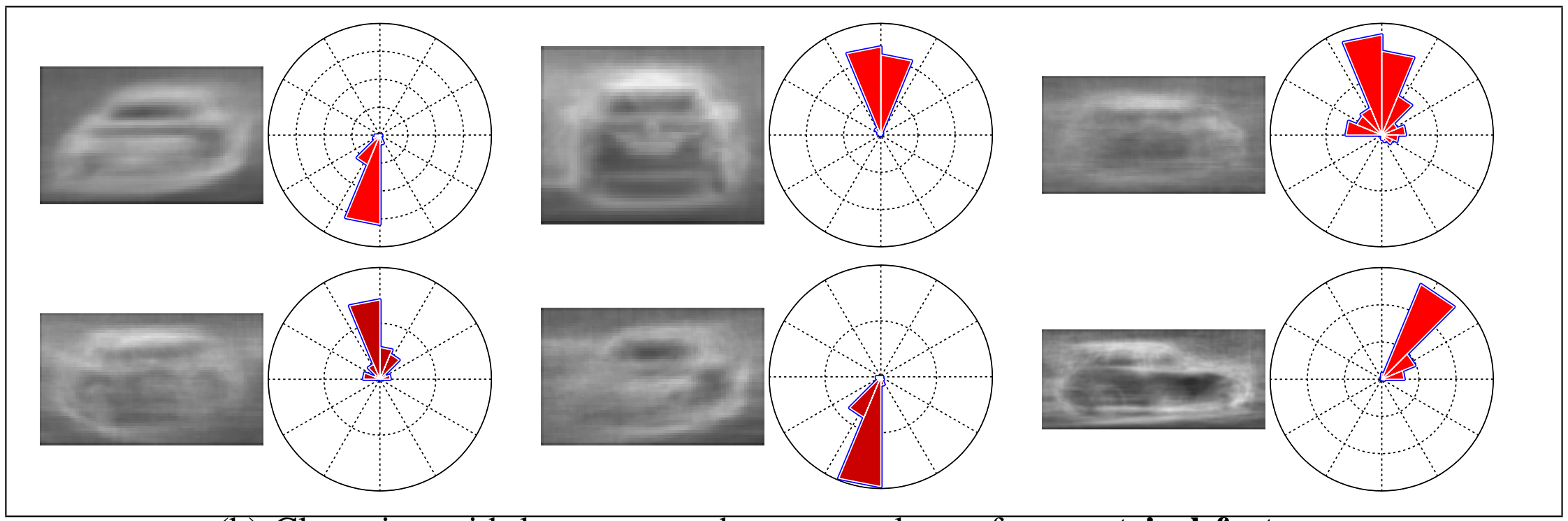

(b) Clustering with k-means on the proposed set of **geometrical features**.

Fig. 5: Clustering to 6 clusters with two types of features. (a) Visual features (color, gradient magnitude, and gradient orientation features at 6 bins) vs. (b) k-means on geometrical features. Rose plots show the orientation distribution of the samples in each centroid. Color shows a percentage of occluded samples, varying from light red (no occlusion) to dark (occlusion). Note how clustering based one visual features provides clusters containing samples at multiple orientations and no clear separation in occlusion level.

over $[0-50\%]$ and $[51\%-100\%]$ for each orientation bin.

**Strategy 2**: To account for all the variables that influence appearance variations (such as occluder statistics, truncation, etc. see Section III-B), we may cluster over 3D geometry features directly, using k-means or spectral clustering (**SC**), which we found to work well (see Fig. 5).

**Strategy 3**: 3D geometry features may be used in order to initialize a visual subcategorization routine, such as LSVM or the framework in [6]. Fusion of the two types of features, visual and geometric, is important as a partition in the geometrical space may not be correlated with a partition in the visual space. Nonetheless, this scheme produced minor improvements on the final detection results in our experiments.

**Strategy 4**: Clustering of visual features only with no geometrical data, as described in III-A.

**Comments on unsupervised or weakly-supervised clustering methods**: k-means and spectral clustering were used for unsupervised clustering. In our implementation of spectral clustering, a Gaussian kernel is employed as a similarity function between two samples $\mathbf{x}_i$ and $\mathbf{x}_j$, $W_{ij} = \exp \frac{||\mathbf{x}_i - \mathbf{x}_j||^2}{2\sigma^2}$. We then compute the normalized graph Laplacian, $L = I - D^{-\frac{1}{2}} W D^{-\frac{1}{2}}$, where $D$ is the diagonal degree matrix [11]. Next, k-means is run on the $L_2$ normalized matrix of eigenvectors of $L$. These two clustering techniques will be compared against the discriminative subcategorization framework of [6] (referred to as **DSC**) both with visual and geometrical features. DSC employs a weakly-supervised framework for obtaining cluster labels with the presence of negative samples. As in [6], our experiments showed DSC to be superior to Latent SVM in prevention of degenerate clusters and overall cluster purity. DSC utilizes a block coordinate gradient-descent alternating between optimization of the SVM parameters and the cluster labels. Different initialization schemes for DSC will be studied. Generally, we found that the procedure of first training a linear SVM on the positive and negative instances to obtain a weight vector $\mathbf{w}$, and clustering the residual vectors after projection on $\mathbf{w}$ using $\mathbf{x} - \frac{1}{||\mathbf{w}||}(\mathbf{w}^T\mathbf{x})\mathbf{w}$ improved the final clustering quality. For negative samples, we use three iterations of hard negative mining.

## IV. Detection Framework

AdaBoost [7] is learned using depth-2 decision trees as weak classifiers. Detection at multiple scales is handled using approximation of features at nearby scales for speed, as in [37]. When computing the 10 feature types, the original image layout is preserved. These are processed by local summation in $4\times4$ blocks, producing a compact descriptor of size $w{\cdot}h{\cdot}10/16$ for a window of size $w \times h$. Unlike common Viola-Jones style techniques, feature classification only involves a value lookup, and not sums over rectangular regions computed with integral image sums. For more detail, see [7]. The approach has been previously applied to diverse tasks such as sign recognition [38] and face detection [39].

**Training parameters**: In all of the experiments, training one component involves several iterations of hard mining of negative instances, with the first round sampling 5000 random negative samples, followed by three additional stages of training with collected hard negatives. In each round, instances belonging to the other subcategories are excluded as negatives. This exclusion takes place according to an overlap threshold (where overlap($b1$,$b2$) $= \frac{area(b1 \cap b2)}{area(b1 \cup b2)}$, for bounding boxes $b1, b2$). Out of the values of $\{0.1, 0.15, 0.2, \ldots, 0.5\}$, $0.3$ was shown to work best for exclusion.

**Pooling detectors**: Given a test image, the trained models for each subcategory are all evaluated. Overlapping detections are merged using a greedy non-maximum suppression (NMS) procedure; once a bounding box is suppressed by the overlap criterion, it can no longer suppress weaker detections. We experimented with two NMS schemes: one using the PASCAL overlap criteria of intersection-over-union, and a second scheme where the union denominator is replaced by the minimum area of the two bounding boxes [37]. Best results were shown with the classical PASCAL NMS scheme, and NMS overlap threshold of $0.3$. Calibration of the models' scores by linearly rescaling to $[0, 1]$ range using the maximum/minimum scores on the validation set before performing NMS is evaluated as well.

**Multiresolution models**: Our experiments will demonstrate that learning multiresolution models significantly improves detection performance. Appearance variation due to the distance of the object from the camera can also be thought of as a subcategorization feature, and discriminative detail may be lost when using a single resolution model [40]. In traditional sliding window object detection, one template model is learned and a feature pyramid is extracted (i.e. by downsampling the original image) for handling detection at multiple scales. An alternative approach was studied in [8], [10], which proposed learning a template pyramid as opposed to a feature pyramid due to speed/memory considerations with a small gain in detection performance. Unfortunately, learning models even just for 5 scales (as in [8]) is very costly, especially due to having to learn multiple components per scale. Therefore, a hybrid approach is studied. In the hybrid approach, templates are learned at multiple resolutions and applied over a feature pyramid. It will be demonstrated how just one or two additional higher resolution models account for most of the performance gain, with a significantly lower cost in training time when compared to a full template pyramid approach.

## V. Orientation Estimation

Due to the close relationship between the subcategorization and vehicle orientation, this immediately motivates the study of orientation estimation. In particular, we care about the relationship between the number of subcategories and orientation estimation accuracy. Furthermore, the impact that the different strategies have on orientation estimation is also of interest.

Possible issues with a large number of subcategory models include resolving scores distribution of nearby orientation models as well as opposite orientations (models with $\pi$ difference in orientation commonly spike in score together). For instance, rear and front instances would sometimes get mixed, as well as left and right orientations. This is intuitive, but requires a more careful analysis of the scores output. Therefore, two approaches were considered for performing the final orientation estimation, one is using classification and one using regression. For regression, we use a L2-regularized L2-loss support vector regression [41]. For classification, we use a Crammer and Singer multiclass SVM [42]. In the latter, a weight $w$ is learned for each class, and these weights are optimized as a whole. Both are used with a linear kernel.

In order to compare among all clustering methods, a mapping is learned from the detectors' scores at sufficient spatial proximity to a 3D orientation value. Given the set of detections in an image, $D$, we construct a feature vector for each detection box as following. Each detection is defined by a bounding box $B$, a score $s$, and the associated model $k$, so that $(B, s, k) \in D$. First, NMS is performed in order to produce sparse detection boxes and fixing detection performance to the one in Section IV. Consequently, for leveraging context from other subcategory models in the final orientation estimate, NMS is performed on each model individually and a feature vector is constructed using the maximum score of each detector that has a higher overlap than 0.5 with the given post-NMS detection. Therefore, $k$ models produce a $k$ dimensional feature vector of scores, $(s_1, ..., s_K)$. This approach provides a general orientation estimation solution independent of the subcategorization technique.

In training, the models are evaluated on the annotated training images, so that each true positive contributes a training sample for the orientation estimation model. The scores are linearly normalized before inputting to the SVM.

Orientation estimation is evaluated using the orientation similarity metric proposed in [31],

$$s(r) = \frac{1}{|D(r)|} \sum_{i \in D(r)} \frac{1 + cos\Delta_\theta^i}{2} \delta_i \qquad (1)$$

where, for a given recall rate $r$, $D(r)$ is the set of object detections and $\Delta_i$ is the angle difference between estimated and ground truth orientation. $\delta_i$ is set to 1 if detection $i$ has been assigned to a ground truth bounding box and 0 otherwise.

## VI. Experimental Settings

For evaluation of the proposed framework, the KITTI dataset is used [31]. Three evaluation methods were suggested in [31], 'easy', 'moderate', and 'hard' with increasing occlusion and truncation and decreasing minimum object size. There are 7481 training images (with over 20,000 vehicle instances), which were split in half to produce a training and a validation dataset. All the experiments employed a 70% overlap requirement in order for a detection to count as a true positive, and are performed by testing with 'moderate' test settings. Although the official performance metric is average precision (AP), we also plot receiver operating characteristic (ROC) common in detection. In parenthesis for all ROC plots the detection rate (recall) at $10^{-1}$ false positives per image (FPPI) rate is shown, and for precision-recall curves the area under the curve (higher is better in both cases).

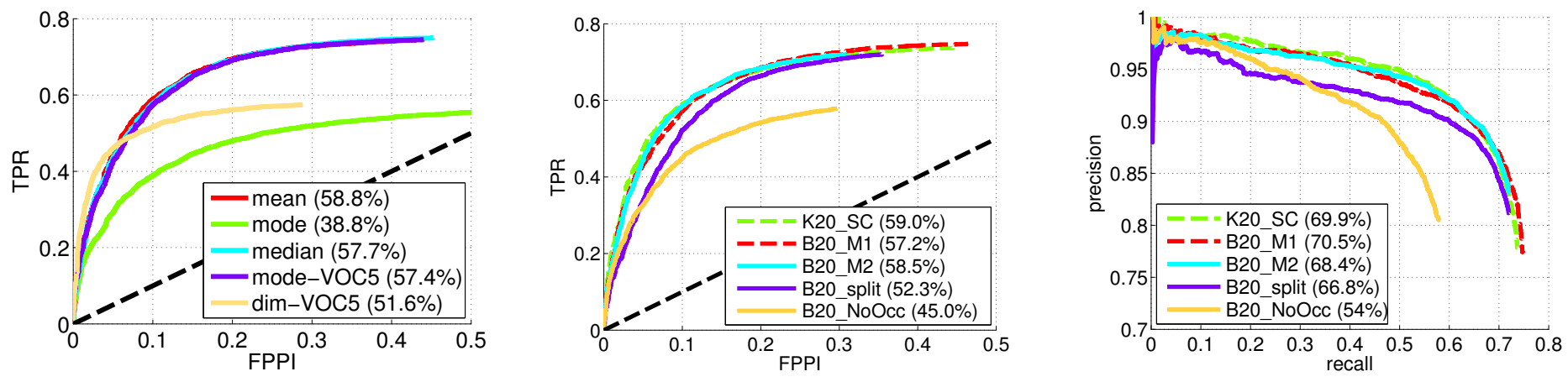

(a) (b) (c)

Fig. 6: (a) Given a clustering assignment, how should model dimensions be determined? (b) and (c) analyze strategy 1 in which a bin parameter in orientation space (**B**) and occlusion level space (**M**) is varied. For $M > 2$, no improvement was gained. The results are compared with strategy 2, spectral clustering on geometrical features ($K$20_SC or **SC (Geo)**) which is the best performing.

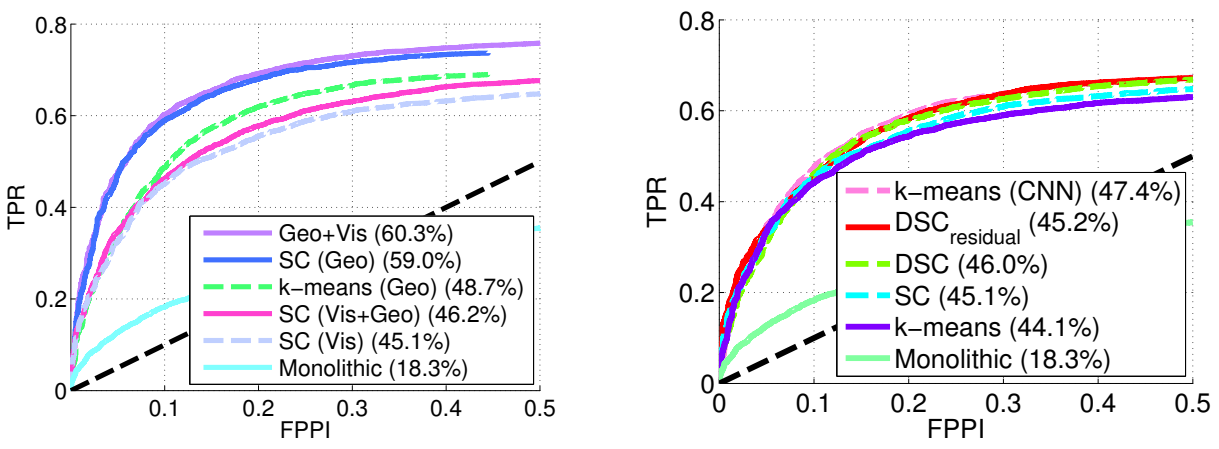

(a) (b)

Fig. 7: Results for $K = 20$ subcategories. (a) Analysis of the clustering techniques and different features (strategies 2 and 4), see Section VII for more detail. Spectral clustering (**SC**) with geometrical (**Geo**) features is shown to work well, better than purely visual (**Vis**) subcategorization. (b) Strategy 4; Clustering analysis for only visual features defined in Section III.

## VII. Experimental Evaluation

For clarification, it is pointed out that throughout the experiments the letter $B$ refers to the number of uniform 3D orientation bins, $M$ refers to occlusion level bins, and $K$ is used for the unsupervised or weakly-supervised clustering experiments such as spectral clustering (**SC**). $K20_SC$ refers to strategy 2, where the geometrical features are clustered to 20 components, and is the same as **SC (Geo)**.

**Model parameters**: The results for this analysis are shown in Fig. 6(a). For obtaining the model dimensions of each cluster, several options were considered. One may determine an aspect ratio for each cluster using the mean, mode, or median of the samples' aspect ratios. In these approaches, one of the dimensions is always kept fixed, and the other is derived from the aspect ratio. Careful optimization showed fixing one dimension (width) at 32 pixels worked best (used in all of the experiments). The DPM-VOC version 5 code [1] is commonly used in object detection studies, yet two approaches based on it were shown sub-optimal. In **mode-VOC5**, the aspect ratios of the samples in each cluster were filtered as in the available implementation in the process of obtaining the mode. Then, a base dimension of 32 was used to obtain the dimensions of each component. In **dim-VOC5**, the entire pipeline from [1] was used, which determines model dimensions by picking the 20th percentile area (as opposed to fixing one dimension at 32). Model padding was also grid optimized, and $1/8$ of the model size in each dimension (width and height) was used for padding. Note that a monolithic classifier runs at ~12.5 fps on a CPU on full resolution images of size $1242 \times 375$.

**Object subcategorization strategies**: Here we study clustering and feature combinations. First, strategy 1 is evaluated in Fig. 6, where it is shown that a good approach is simply to set $M = 1$ for occlusion handling under moderate test settings. Little benefit was made by learning occlusion separate models as shown for $M = 2$. Performing **split**, where fully visible samples are separated from samples with any kind of occlusion, also under-performed the $M = 1$ binning. $M = 1$ corresponds to keeping both occluded and non-occluded samples in the same cluster. These results are in contrast to the results of the study of [5] (possibly due to the different baseline detectors, no parts, etc.) and a common practice in pedestrian detection where occluded samples are excluded [43]. The results were consistent with the same strategy on the 'hard' train/test settings. That is, when training a rigid template component, a much better detection model for heavily occluded vehicle instances was produced when the training cluster was allowed to contain all no-occlusion, partial-occlusion, and heavy-occlusion samples at the same orientation, as opposed to just heavy-occlusion samples or partial- and heavy-occlusion instances. Generally, quantization over the occlusion space resulted in significant reduction to cluster sizes and consequently lower performing models (this also explains why the split scheme performs poorly). Another important observation is that training on 'easy', 'moderate' and 'hard' settings resulted in a detector with different strengths, limitations, and optimal NMS settings. Strategy 2 and SC, which leverage information from the different geometrical features, was also shown to work well among all strategies for a fixed size of $K = 20$.

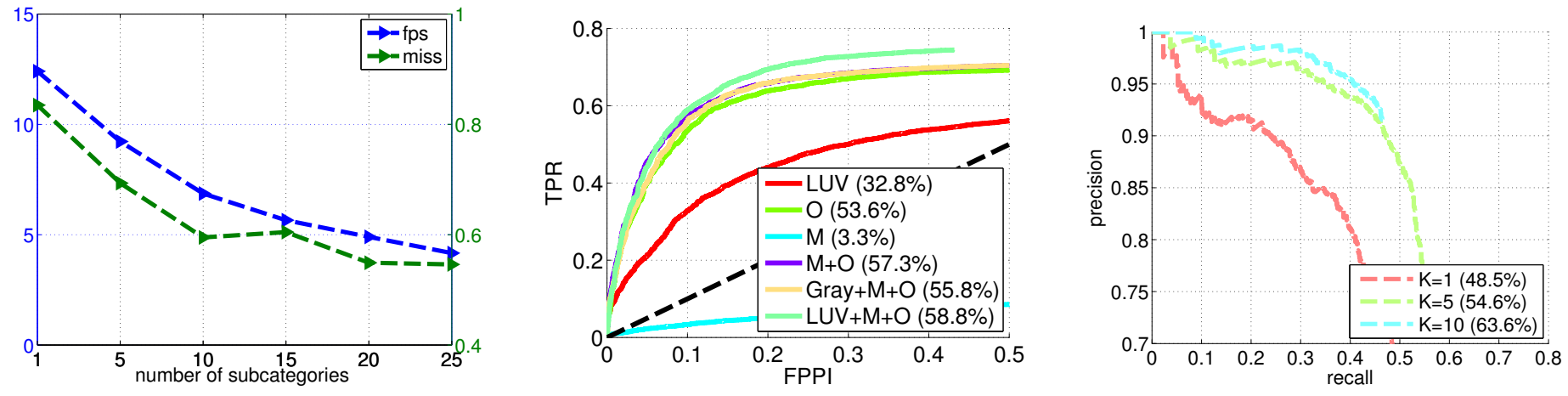

(a) (b) (c)

Fig. 8: (a) Impact of varying the number of subcategories ($K$) on frame-rate of the entire detection pipeline and detection performance (miss rate at $10^{-1}$ FPPI). (b) Impact of the feature types (LUV or grayscale image channels, M-normalized gradient magnitude, O-gradient orientation) on detection performance. (c) Using the proposed framework to learn subcategory models for pedestrian detection favorably impacts performance over a monolithic classifier ($K = 1$).

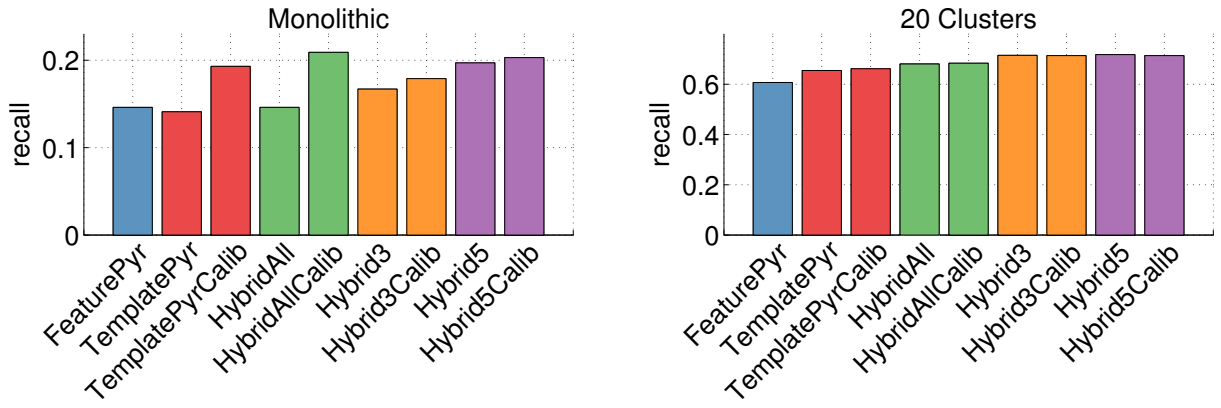

(a) (b)

Fig. 9: Comparative evaluation of varying detection techniques in a one cluster settings (monolithic) and 20 clusters settings. Adding higher resolution components, just at one or two other model scales, results in significant detection performance gains with the hybrid approach. For approaches involving multiple templates, the effect of score calibration is shown.

The remaining strategies are analyzed in Fig. 7. In employing LUV+gradient features, DSC was shown to produce better detection results when compared to SC or just k-means. The original implementation of DSC [6] was modified to use SC for initialization as opposed to k-means, which provided improved detection results. As no benefit was shown by providing DSC an initialization based on ground truth orientation or occlusion labeling (strategy 3), it is not compared in Fig. 7. This may be due to the disagreement between the visual and geometrical modalities. Interestingly, for CNN features, k-means worked as well as SC, hence only the curve for k-means is shown. The CNN results are shown for comparative analysis, yet it still significantly outperformed by the geometrical subcategory models.

**Fusion of geometric and visual features**: As aforementioned, DSC initialization using geometry features did not significantly improve the final detection performance. Furthermore, our attempts to leverage the two types of modalities resulted in reduced detection accuracy. Concatenation of visual and geometrical features hindered detection performance as well. The following two approaches have shown a small improvement due to fusion: 1) In the computation of the affinity matrix $W$ before applying SC, the Euclidean similarities were computed separately for visual and geometrical features. A linear combination of these was then taken to produce $W$, with most weight given to the geometrical features. 2) A similar improvement was gained by learning a separate set of $K = 20$ subcategories for each modality and running all 40 models (shown as **Geo+Vis** in Fig. 7(b)). The gains were small at a high cost of additional subcategory models, hence SC and geometrical features were used in most of the analysis.

**Number of subcategories**: As shown in Fig. 8, $K = 20$ provided a good choice. Even with 20 models, the method detects at $\sim$5 fps on a CPU on full resolution images.

**Conclusion for subcategorization**: Either uniform orientation binning as in strategy 1 with $M = 1$ or clustering geometrical features as in strategy 2 work well. Visual categorization, even with high quality CNN features or state-of-the-art clustering techniques, was shown to be a difficult task.

**Hybrid multiresolution detection**: In Fig. 9, the impact of multiresolution models is studied both for a monolithic classifier and a 20 subcategories detector. Five approaches are compared: the traditional feature pyramid with one template approach (**FeaturePyr**), the template pyramid approach from [8] (**TemplatePyr**), the hybrid approach at all possible scales (**HybridAll**), and the hybrid approach corresponding to learning a total of three resolution models (**Hybrid3**) and five resolution models (**Hybrid5**) corresponding to model width of $\{w, \frac{5}{4}w, \frac{3}{2}w\}$ and $\{w, \frac{5}{4}w, \frac{3}{2}w, 2w, 3w\}$, respectively ($w = 32$ pixels works best). Note how components at different resolutions greatly improve recall rate at $10^{-1}$ FPPI. Furthermore, simply adding one or two resolution components at a slightly higher resolution than the 32 pixels baseline with the hybrid approach is shown to account for most of the performance gain. This results in significant reduction in training time over the template pyramid baseline, without loss in detection performance. This is consistent with the 20 subcategories detector case as well, where the hybrid3 approach performs best. Learning more than 3 resolution components is shown to be not necessary and even harmful. Calibration of each model output scores to $[0-1]$ range is shown to be helpful under settings with many resolution models, but with 'the few that matter' has little impact. Although run time speed decreases with each resolution addition, the improvement in

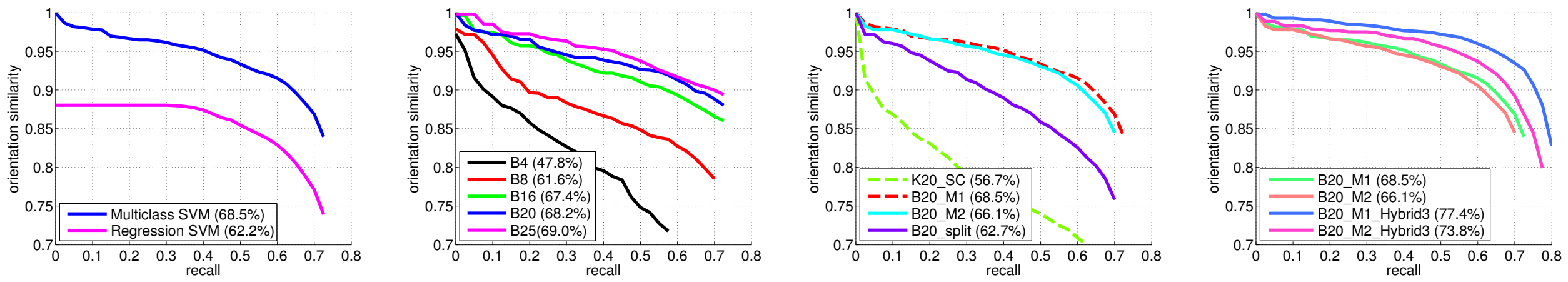

(a) (b) (c) (d)

Fig. 10: Impact of subcategorization on vehicle orientation estimation. (a) Classification with a multiclass SVM vs. support vector regression. (b) Impact of number of orientation bins (at occlusion quantization $M = 1$) on orientation similarity results. (c) Different occlusion handling techniques and their effect on orientation estimation. Although K20_SC provided the best detection results, the unsupervised framework comes at a cost for orientation estimation. (d) The presence of both occluded samples and non-occluded samples in the same cluster (as in $M = 1$) provide a clear advantage, especially when learning multiresolution models.

TABLE II: Evaluation on the KITTI testing benchmark. (a) Area under the precision-recall curve with varying test settings for the detection task. The top three methods are shown in bold. (b) Area under the orientation similarity-recall curve for detection and orientation estimation evaluation. DPM-based methods employ version-4 of the available implementation unless stated as **V5**.

(a) Car detection

| **Method** | **Easy** (%) | **Moderate** (%) | **Hard** (%) |
|---|---|---|---|
| Regionlets [44], [45] | **84.27** | **75.58** | **59.20** |
| **SubCat (Ours)** | **81.94** | **66.32** | 51.10 |
| AOG [17] | **80.26** | **67.03** | **55.60** |
| SpCov_ACF [46] | 78.67 | 58.19 | 44.80 |
| DPM (V5) [17] | 77.24 | 56.02 | 43.14 |
| SpCov [46] | 76.53 | 62.29 | 48.00 |
| OC-DPM [5] | 74.94 | 65.95 | **53.86** |
| DPM-C8B1 [47] | 74.33 | 60.99 | 47.16 |
| LSVM-MDPM-sv [1] | 68.02 | 56.48 | 44.18 |
| LSVM-MDPM-us [1] | 66.53 | 55.42 | 41.04 |
| ACF [7] | 55.89 | 54.74 | 42.98 |
| mBoW [48] | 36.02 | 23.76 | 18.44 |

(b) Car detection and orientation estimation

| **Method** | **Easy** (%) | **Moderate** (%) | **Hard** (%) |
|---|---|---|---|
| **SubCat (Ours)** | **80.92** | **64.94** | 50.03 |
| OC-DPM [5] | 73.50 | 64.42 | **52.40** |
| LSVM-MDPM-sv [1] | 67.27 | 55.77 | 43.59 |
| DPM-C8B1 [47] | 59.51 | 50.32 | 39.22 |
| AOG [17] | 44.41 | 36.87 | 30.29 |

performance is significant. Furthermore, the overall method is still significantly faster than comparable methods, such as the DPM (even with multiresolution models). Nonetheless, there many possible speedups which could explore the redundancy among the models and detection at different scales.

**Generalization to pedestrian detection**: The proposed framework can be applied for other domains of object detection. For pedestrian detection, we follow the same framework, with minor changes to the model dimensions (64 pixels height works well, with additional lower resolution component at 32 pixels). Results are shown in Fig. 8. For this evaluation, an overlap threshold of 50% is used.

**Orientation estimation results**: As shown in Fig. 10(a), the multiclass SVM produced significantly better orientation estimation compared to support vector regression which outputs a continuous value. The number of orientation bins is analyzed in Fig. 10(b), and a plateau is seen after $B = 25$. For these experiments, we set $M = 1$. The analysis for occlusion level binning analysis in vehicle detection is consistent with the results in Fig. 10(d), where $M = 1$ is shown to work well both in the single resolution and multiple resolution components (hybrid) case.

**Comparison with state-of-the-art**: Table II shows the detection and orientation estimation results on the KITTI testing dataset. We also refer the reader to the online evaluation board at http://www.cvlibs.net/datasets/kitti. Since the detection task contains many entries, the three top methods in each evaluation category are bolded. Interestingly, some techniques perform well on detection (AOG) but poorly on orientation estimation. Because the different approaches employ different computational environments, no explicit comparison in terms of speed can be made. Nonetheless, our submission provides a speedup of about a factor of 10-30 over reported speeds of varying DPM-based approaches. Remarkably, the fastest techniques which employ the same baseline detection framework as ours (ACF) do not nearly perform at the same level, even with richer features [46]. Our approach is shown to significantly improve detection performance over an out-of-the-box multi-component ACF-based submission by a large margin of 26%, 11%, and 8% in 'easy', 'moderate', and 'hard' test settings, respectively. This in turn brings the ACF approach from one of the lowest-performing approach to one of the top-performing. Other current state-of-the-art methods employ features, such as CNN, Local Binary Patterns (LBP), and Covariance features. The improvement from adding such features is orthogonal to our approach, which only utilizes HOG+LUV features.

## VIII. Concluding Remarks

In this paper, the role of learning appearance patterns of an object type (vehicles) for detection and orientation estimation was studied. An extensive set of experiments demonstrated that when training rigid templates with AdaBoost, geometrical subcategorization resulted in improved detector performance. Further study of the fusion of visual and geometrical modalities is left for future work. Forming clusters corresponding to occlusion levels resulted in good detection only when clusters were kept large by incorporating non-occluded samples (i.e. learning a model on 'moderate' or 'hard' settings, as opposed to just on 'hard' instances). Learning models at multiple resolution was shown to significantly improve detection/orientation estimation performance. A large drop in performance was observed when using the more strict 70% overlap evaluation threshold as opposed to the common 50% (also shown in [47]), indicating better localization is required. This could be addressed using regression approaches, as in [44], [49], although heavy occlusion is still the main challenge in detection (Fig. 11). Further improvements can be made by incorporating scene information [50]–[53] (see Fig. 12). The fast detection approach of [9] may be used for further speedups. Finally, we would like to study application of the framework to other domains, such as hand detection [54].

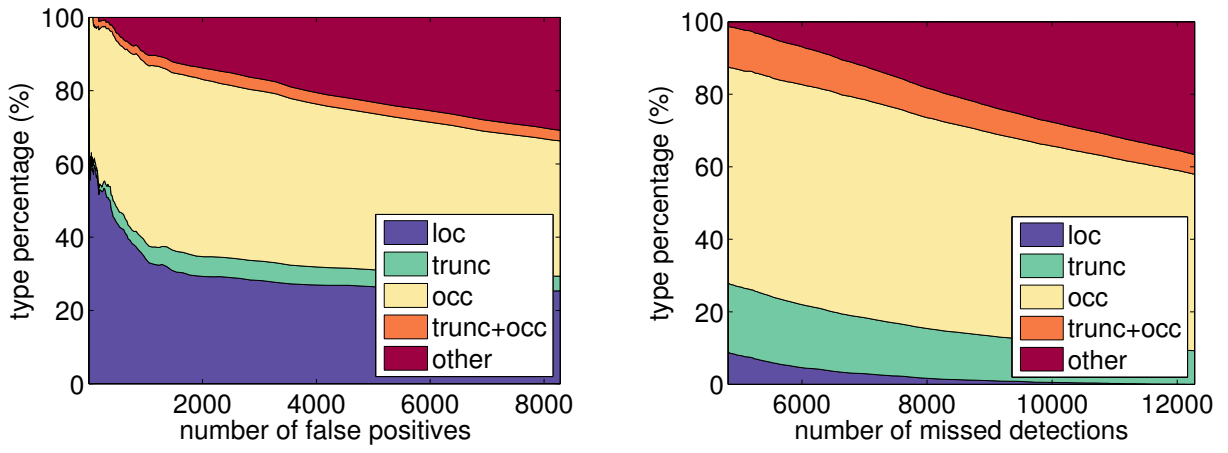

(a) (b)

Fig. 11: The distribution of types of false positives and missed detections. The x-axis is associated with decreasing score threshold. False positive or missed detection boxes may occur due to poor localization (loc, overlap requirement of 0.1 or more), occlusion (occ), truncation (trunc), both occlusion and truncation (trunc+occ), or another reason (other). Vehicle detection under heavy occlusion is still a challenge.

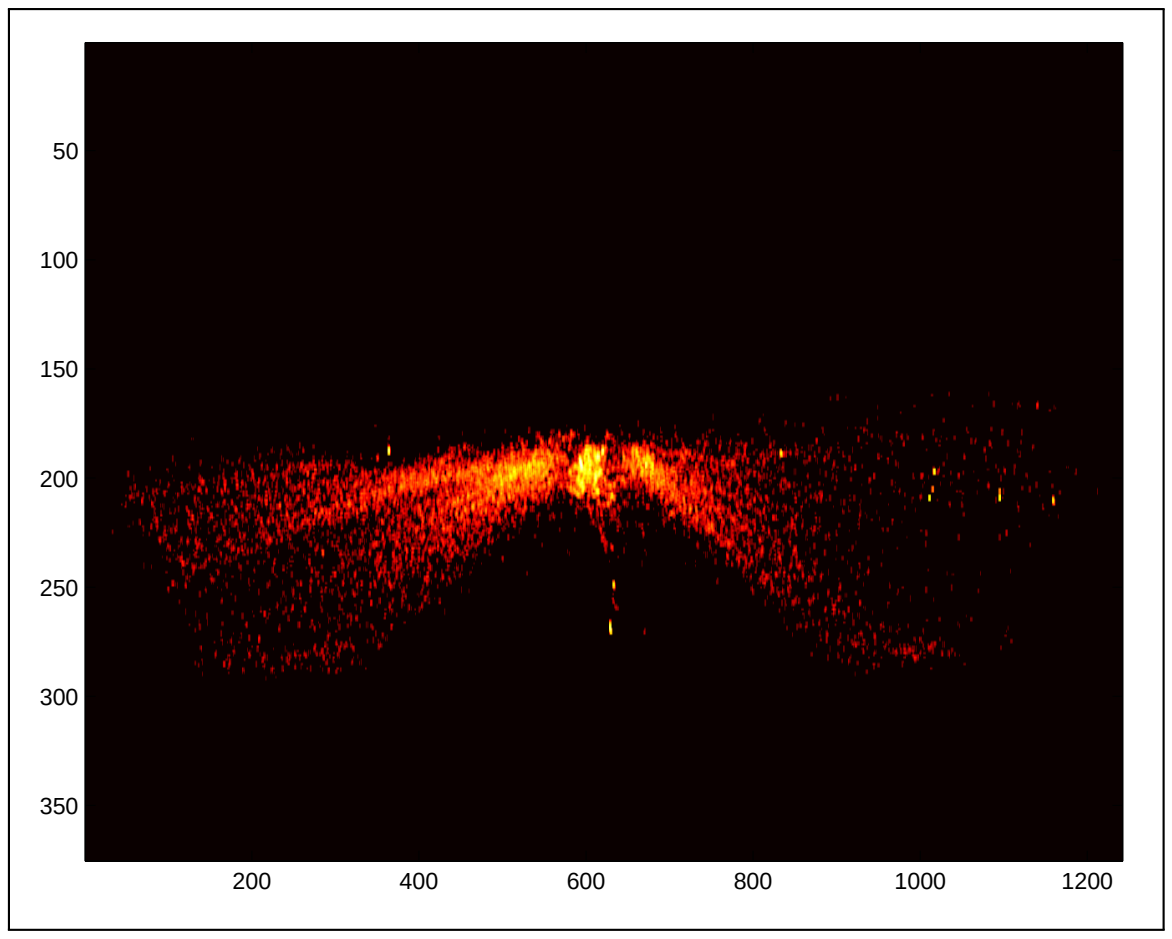

Fig. 12: Center location distribution (log normalized) obtained from ground truth. Scene information can be leveraged for improved detection performance and possible speedups, as in [51]. This is left for future work.

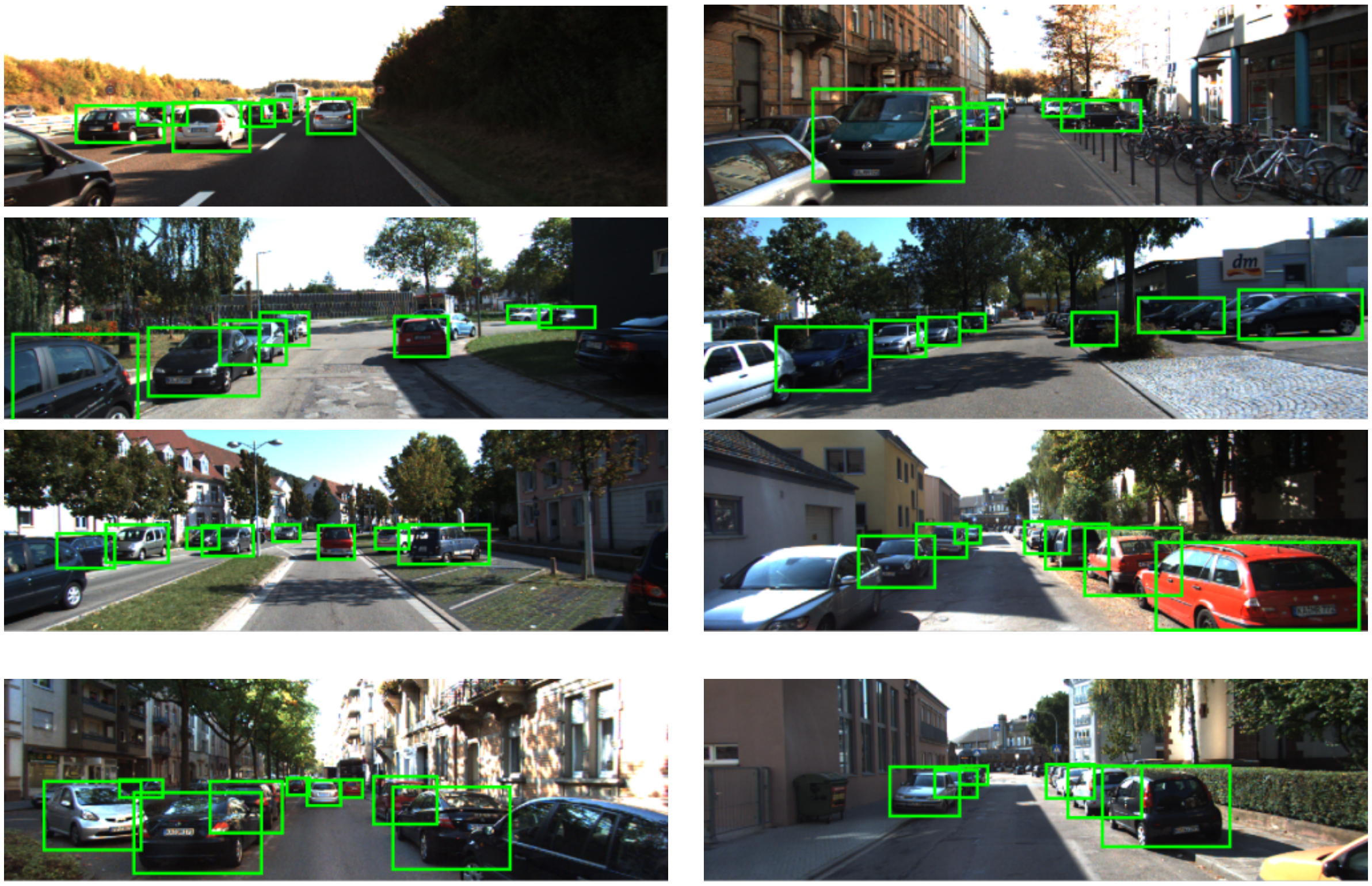

Fig. 13: Detection results on the KITTI dataset. Future work should improve occlusion handling, localization tightness, and truncated vehicle detection.

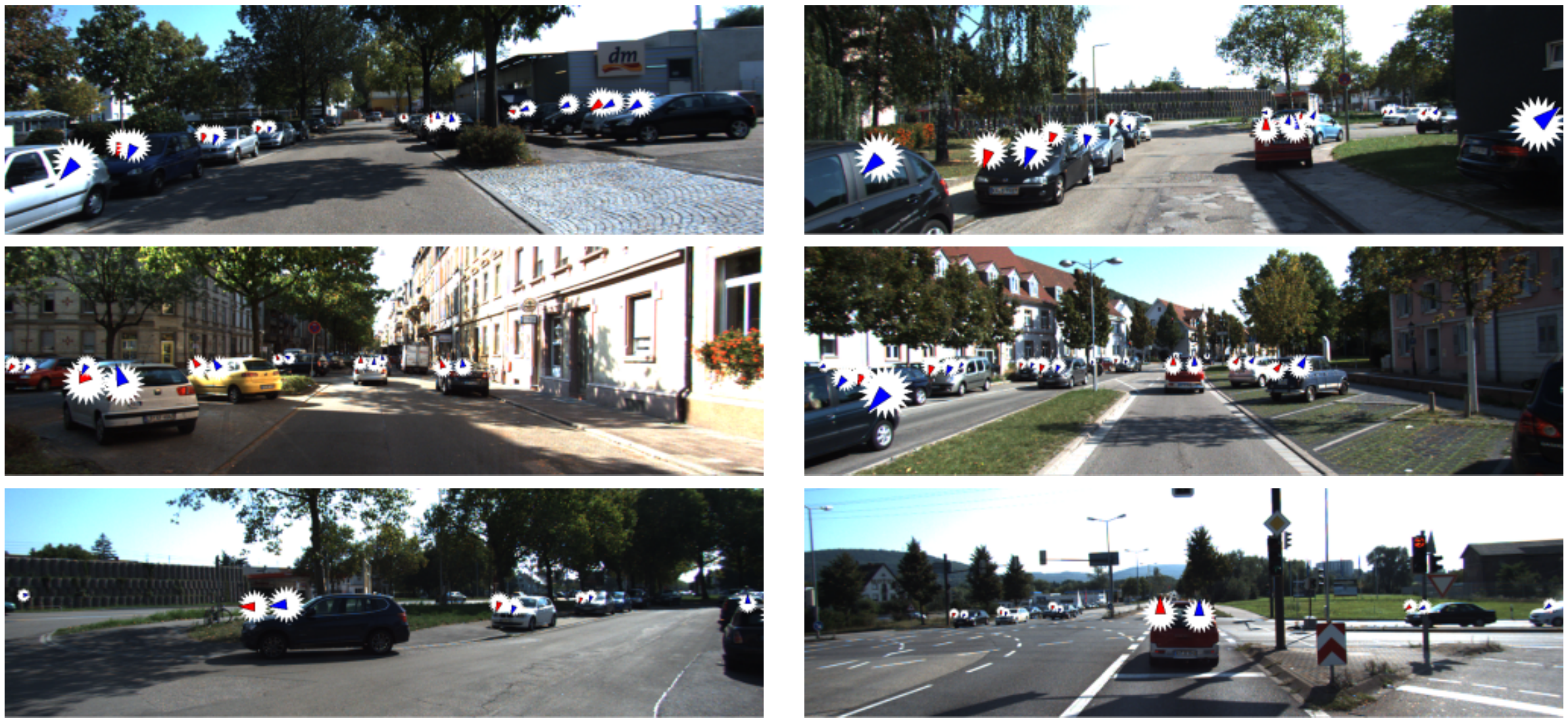

Fig. 14: Orientation estimation results on the KITTI dataset. In red is the estimated orientation and in blue the ground truth orientation.

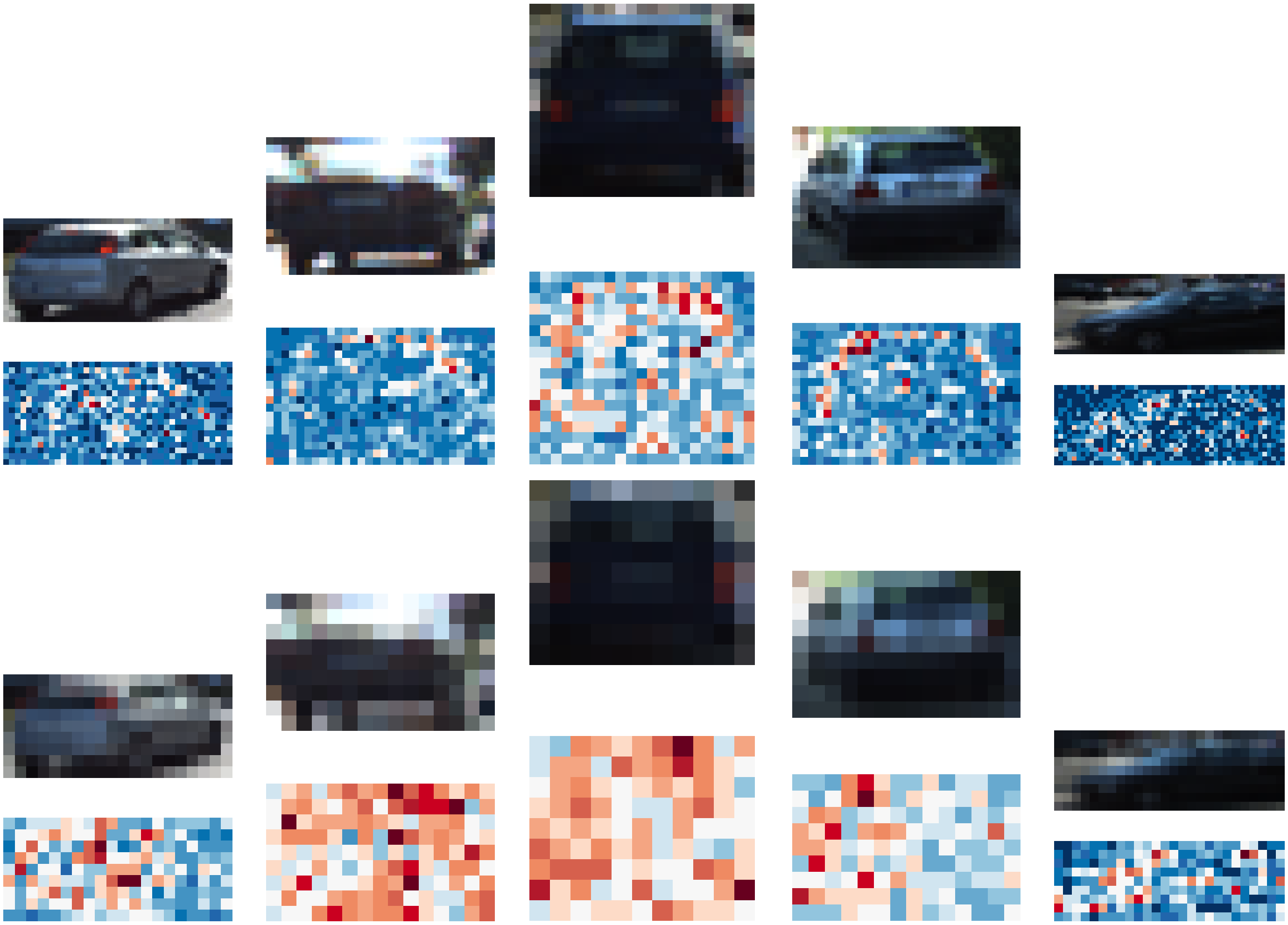

Fig. 15: Learning components at multiple resolutions (visualized here for widths 64 and 32 pixels in the top and bottom row, respectively) with the hybrid multiresolution approach results in significant performance improvement. Warmer colors correspond to feature locations with more attention given by the classifier.

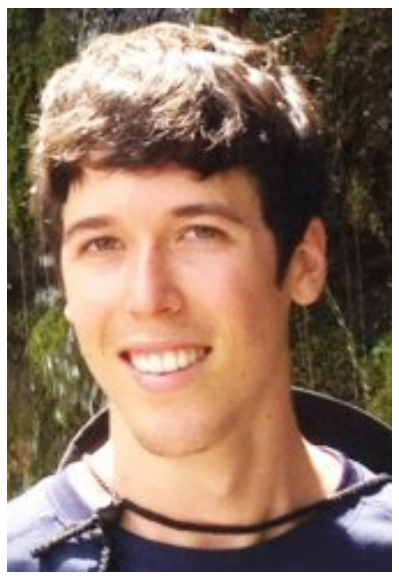

**Eshed Ohn-Bar** received his M.S. degree in electrical engineering from the University of California, San Diego (UCSD) in 2013 and is currently pursuing a Ph.D. with a focus on signal and image processing at UCSD. His research interests include computer vision, object detection, multi-modal activity recognition, intelligent vehicles, and driver assistance and safety systems.

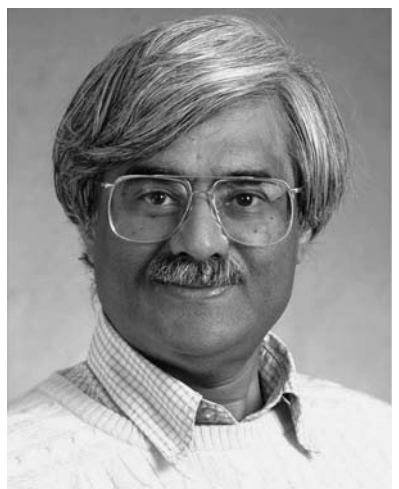

**Mohan Manubhai Trivedi** received the B.E. (with honors) degree in electronics from Birla Institute of Technology and Science, Pilani, India, in 1974 and the M.S. and Ph.D. degrees in electrical engineering from Utah State University, Logan, UT, USA, in 1976 and 1979, respectively. He is currently a Professor of electrical and computer engineering and he is the Founding Director of the Computer Vision and Robotics Research Laboratory, University of California San Diego (UCSD), La Jolla, CA, USA. He has also established the Laboratory for Intelligent and Safe Automobiles, Computer Vision and Robotics Research Laboratory, UCSD, where he and his team are currently pursuing research in machine and human perception, machine learning, human-centered multimodal interfaces, intelligent transportation, driver assistance, active safety systems and Naturalistic Driving Study (NDS) analytics. His team has played key roles in several major research initiatives. These include developing an autonomous robotic team for Shinkansen tracks, a human-centered vehicle collision avoidance system, a vision-based passenger protection system for smart airbag deployment, and lane/turn/merge intent prediction modules for advanced driver assistance. He regularly serves as a Consultant to industry and government agencies in the United States, Europe, and Asia. He has given over 70 Keynote/Plenary talks at major conferences. Prof. Trivedi is a Fellow of the International Association of Pattern Recognition (for contributions to vision systems for situational awareness and human-centered vehicle safety) and the Society for Optical Engineering (for contributions to the field of optical engineering). He received the IEEE Intelligent Transportation Systems Societys highest honor, Outstanding Research Award in 2013, the Pioneer Award (Technical Activities) and the Meritorious Service Award of the IEEE Computer Society, and the Distinguished Alumni Award from Utah State University, Logan, UT, USA. He is a co-author of a number of papers winning Best Papers awards. Two of his students were awarded Best Dissertation Awards by the IEEE ITS Society (Dr. Shinko Cheng 2008 and Dr. Brendan Morris 2010) and his advisee Dr. Anup Doshis dissertation judged among the five finalists in the 2011 by the Western (USA and Canada) Association of Graduate Schools. He serves on the Board of Governors of IEEE ITS Society and on the Editorial advisory board of the IEEE Trans on Intelligent Transportation Systems.